\documentclass{article}
\usepackage{ijcai26}

\usepackage{times}
\usepackage{url}
\usepackage[hidelinks]{hyperref}
\usepackage[utf8]{inputenc}
\usepackage[small]{caption}
\usepackage{graphicx}
\usepackage{booktabs}
\usepackage{amsmath,amssymb}

\title{Measuring the Tokenization Premium: A Cost Audit for Underserved Language Communities}

\author{
Avijit Roy$^{1}$ \and
Proma Roy$^{2}$ \And
Hrishitva Patel$^{3}$\\
\affiliations
$^{1}$John Jay College of Criminal Justice, City University of New York, USA\\
$^{2}$The City College of New York, City University of New York, USA\\
$^{3}$University of Texas at San Antonio, USA\\
\emails
aroy@jjay.cuny.edu,
proma.roy33@stu-mail.ccny.cuny.edu,
hrishitva.patel@utsa.edu
}

\begin{document}

\maketitle

\begin{abstract}
Large language models are increasingly deployed as general-purpose educational and technical assistance systems, but their basic infrastructure does not treat languages equally. One under-examined source of disparity is tokenization: semantically equivalent content can require substantially different token counts across languages, affecting API cost, latency, and usable context length before a model is even invoked. We introduce the Tokenization Equity Audit (TEA), a reproducible benchmark for measuring tokenization premiums in technical tutoring content. TEA evaluates three widely used tokenizers, GPT-4o's \texttt{o200k\_base}, Qwen2.5-7B, and Mistral-7B, on a 120-item Python debugging corpus translated from English into Bengali, Hindi, Arabic, Tamil, and Yoruba. Bengali and Hindi are the primary validated cases; the remaining languages provide exploratory cross-script and cross-family comparisons. Across this corpus, Bengali requires 1.56$\times$ as many GPT-4o tokens as English, reducing a nominal 128k-token context window to an effective 82k-token equivalent for the same semantic content. On Qwen2.5 and Mistral tokenizers, Bengali reaches 4.5$\times$ the English token count. Yoruba, despite using Latin script, shows the highest GPT-4o penalty at 2.37$\times$, indicating that tokenization inequity is not reducible to script family alone. We demonstrate that tokenization creates measurable economic and functional barriers that must be addressed as an equity-relevant infrastructure layer for underserved language communities, especially where educational systems depend on low-cost or offline-capable AI tools.
\end{abstract}

\section{Introduction}

Large language models (LLMs) are often presented as scalable tools for educational support, programming assistance, and multilingual access. Yet access to these systems is not determined only by model accuracy. It is also shaped by infrastructure choices: what data models are trained on, what evaluation tasks are prioritized, which languages are represented in vocabularies, and how deployment costs are calculated. For underserved language communities, these choices can create practical disadvantages before any downstream model response is generated.

This paper focuses on tokenization, the process that maps raw text into the token sequences consumed by language models. Prior work has shown that tokenizer design can introduce unequal treatment across languages, including longer tokenized sequences, higher commercial API costs, and reduced usable context windows for non-English users \cite{petrov2023tokenizers,ahia2023cost}. These effects matter directly for low-resource and underserved communities because most commercial LLM services charge by token, and most open-weight LLM deployments are constrained by sequence length, memory, and inference latency.

The problem is especially visible in technical education. Programming tutoring content contains a mixture of natural language, error names, code snippets, identifiers, and instructional explanations. For learners using an LLM-based tutor in a language such as Bengali, the same debugging explanation may consume many more tokens than an English equivalent. The result is not only higher price. It also means shorter effective context windows, faster conversation truncation, and greater pressure on local inference systems designed for low-connectivity settings.

Bengali also presents a useful case because its technical educational ecosystem relies heavily on English-derived programming terminology and mixed-script instructional practice. Recent Bangla language modeling work has documented the need for dedicated Bangla LLMs and code-specific resources \cite{raihan2025tigerllm,raihan2025tigercoder}. Separately, prior infrastructure analysis has argued that Bengali educational AI is affected by a compound set of barriers, including limited web presence, training data imbalance, tokenization penalties, and connectivity constraints \cite{roy2026structural}. TEA operationalizes one measurable component of that broader infrastructure problem: the tokenization premium imposed on technical tutoring content.

We make three contributions:

\begin{enumerate}
    \item We introduce TEA, a small but reproducible benchmark for auditing tokenization premiums in Python debugging and tutoring content across six languages.
    \item We compare three tokenizer families, GPT-4o \texttt{o200k\_base}, Qwen2.5-7B, and Mistral-7B, using token fertility ratio, effective context window, and illustrative API cost premium metrics.
    \item We show that tokenization penalties are large enough to have practical implications for underserved-language educational AI, including Bengali offline tutoring systems and commercial API-based support tools.
\end{enumerate}

The goal of TEA is not to claim comprehensive coverage of all languages or all educational domains. Instead, it provides a targeted measurement protocol for a concrete use case: beginner programming help in languages that are underrepresented in mainstream LLM infrastructure.

\section{Related Work}

\subsection{Tokenizer Inequity}

Tokenizers are often treated as preprocessing tools, but recent work shows that they can shape fairness, efficiency, and downstream accessibility. Petrov et al. demonstrate that language model tokenizers can produce drastically different sequence lengths for semantically equivalent text across languages, with direct implications for cost, latency, and context capacity \cite{petrov2023tokenizers}. Ahia et al. analyze commercial language model APIs and show that token-based pricing can lead to non-uniform cost burdens across languages \cite{ahia2023cost}. Rust et al.\ further show that tokenizer quality affects multilingual model performance~\cite{rust2021tokenizer}. 
Thakur et al.\ examine structural causes of cross-lingual 
tokenization disparity and argue for rethinking tokenizer 
design for underrepresented languages~\cite{thakur2025tokenizerdesign}.

TEA builds on this literature but narrows the audit to a domain where token count has immediate educational consequences: programming debugging explanations. Rather than measuring general-purpose parallel text, TEA evaluates technical tutoring content that mixes code, English error names, and natural-language explanations. This domain is relevant because programming education is one of the most common uses of LLM-based assistance and because technical vocabulary often behaves differently from general conversational text.

\subsection{Bangla and Low-Resource Technical NLP}

Recent Bangla LLM research has begun to address gaps in general language modeling and code generation. TigerLLM introduces a family of Bangla language models and reports improvements over prior Bangla baselines \cite{raihan2025tigerllm}. TigerCoder extends this direction into Bangla code generation, including Bangla code instruction datasets and MBPP-Bangla \cite{raihan2025tigercoder}. These studies provide important model and dataset contributions, but the infrastructure cost of tokenizing Bangla technical content remains a separate issue.

Prior work on Bengali educational AI has also emphasized that low-resource language problems are not limited to model quality. They include dataset scarcity, instructional language barriers, and deployment constraints in low-connectivity environments \cite{roy2026structural}. TEA complements that argument with a direct measurement of tokenization cost.

\subsection{Offline and Resource-Constrained Educational AI}

Offline-first educational AI systems are motivated by
connectivity and affordability constraints. Locally deployable 
models for low-resource language tutoring can be built with 
parameter-efficient fine-tuning and quantized open-weight models, 
but such approaches do not remove tokenization inefficiency.
They shift the cost from API bills to sequence length, memory pressure, and inference latency. TEA, therefore, treats tokenization as a shared bottleneck across cloud and local deployment models.

\section{The TEA Benchmark}

\subsection{Corpus Design}

The TEA corpus contains 120 English source items drawn from beginner Python programming education. Items are divided into three tiers:

\begin{itemize}
    \item \textbf{Tier 1: short technical phrases} (35 items), such as error names, short diagnostics, and identifiers. Example: ``list index out of range.''
    \item \textbf{Tier 2: short explanations} (50 items), usually one to three sentences explaining a bug or fix.
    \item \textbf{Tier 3: longer tutoring content} (35 items), including conceptual explanations and small code examples.
\end{itemize}

Items were selected to represent common beginner Python debugging scenarios: runtime errors (division by zero, type mismatches), indexing errors (list out of bounds), syntax issues (missing colons, indentation), and correct-code explanations. The selection prioritizes error types frequently encountered in introductory programming courses rather than comprehensive coverage of all Python exceptions.

Each item was translated into Bengali, Hindi, Arabic, Tamil, and Yoruba. Bengali and Hindi are the primary validated languages in this version of the benchmark. Bengali items were reviewed by two Bengali speakers with programming experience, while Hindi items were reviewed by a Hindi-speaking contributor familiar with technical programming terminology and instructional usage. The remaining languages are included as exploratory comparison languages. Their translations were generated with machine translation and manually inspected for obvious semantic or formatting failures, but they should not be interpreted as fully validated pedagogical translations.

Python-specific identifiers, error names, and code tokens, such as \texttt{TypeError}, \texttt{IndexError}, and \texttt{NoneType}, were retained in English across all languages. This reflects common programming practice and avoids mistranslating language-specific symbols or API names. This retention is especially common in longer explanatory and tutorial-style programming content, where English technical terminology frequently coexists with native-language instructional text.

\subsection{Bengali Code-Switching Quality Check}

Technical Bengali often uses code-switching for programming keywords and error names. Code-switching is a standard feature of Bengali programming instruction, not a data quality defect. Treating this as a translation defect would be misleading. To check whether English retention artificially reduces Bengali tokenization penalty, we compute the share of Bengali-script characters in each Bengali item. Items are classified as:

\begin{itemize}
    \item \textbf{Clean:} at least 75\% Bengali-script alphabetic characters,
    \item \textbf{Mixed:} 40\% to 75\% Bengali-script alphabetic characters,
    \item \textbf{English-retained:} below 40\% Bengali-script alphabetic characters.
\end{itemize}

We report results for all Bengali items and for the clean-only subset. This sensitivity analysis checks whether the main Bengali result depends on mixed technical phrasing.

\subsection{Tokenizers}

We evaluate three tokenizers representing common commercial and open-weight deployment settings:

\begin{enumerate}
    \item \textbf{GPT-4o \texttt{o200k\_base}:} the OpenAI tokenizer accessed through \texttt{tiktoken} \cite{tiktoken2026}.
    \item \textbf{Qwen2.5-7B:} a byte-level BPE tokenizer from the Qwen2.5 model family \cite{qwen2025technical}.
    \item \textbf{Mistral-7B-v0.1:} the tokenizer distributed with Mistral-7B-v0.1 \cite{jiang2023mistral,huggingfaceMistralDocs}.
\end{enumerate}

The audit uses tokenizer vocabularies only. No model inference is required for the tokenization measurements.

\begin{figure*}[t]
\centering
\includegraphics[width=0.96\linewidth]{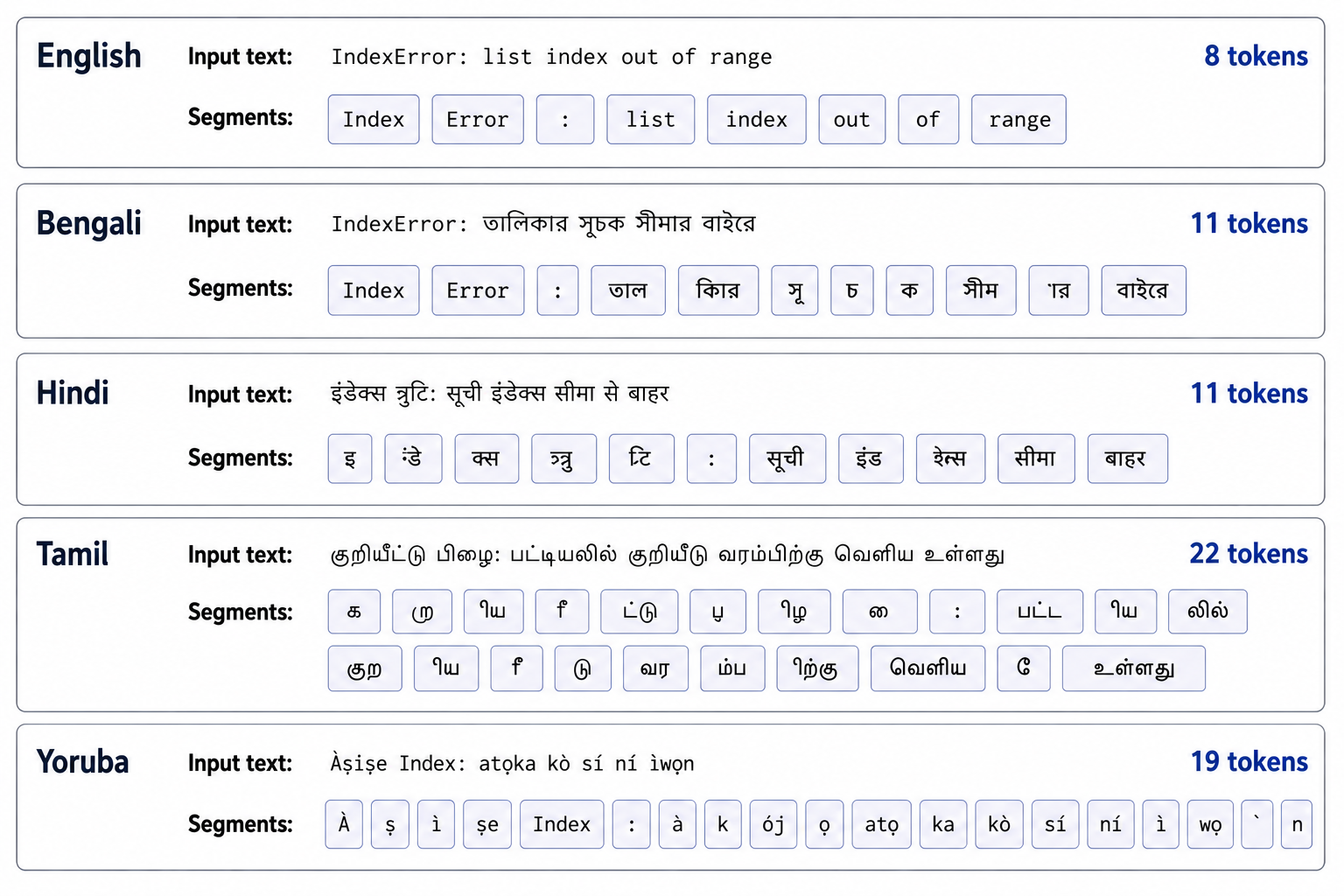}
\caption{Illustrative tokenizer segmentation differences under GPT-4o \texttt{o200k\_base}. English is segmented into relatively compact units, while Bengali, Hindi, Tamil, and Yoruba exhibit greater fragmentation into shorter token pieces. Yoruba is notable because it uses Latin script but still incurs heavy segmentation, demonstrating that tokenization inequity is influenced not only by script family but also by vocabulary coverage, diacritics, and tokenizer training distribution.}
\label{fig:tokenizer_fragmentation}
\end{figure*}

To provide qualitative intuition for tokenization inequity, Figure~\ref{fig:tokenizer_fragmentation} shows example GPT-4o \texttt{o200k\_base} tokenizer segmentations for semantically equivalent debugging phrases across languages. Although illustrative rather than exhaustive, these examples demonstrate how underserved-language text can fragment into substantially more subword units than English, contributing directly to the token fertility ratios reported in Section~5.

\section{Metrics}

\subsection{Token Fertility Ratio}

For each item $i$, language $\ell$, and tokenizer $t$, the token fertility ratio (TFR) is defined as:

\begin{equation}
\mathrm{TFR}_{i,\ell,t} = \frac{\mathrm{tokens}(x_{i,\ell}, t)}{\mathrm{tokens}(x_{i,\mathrm{English}}, t)}.
\end{equation}

A TFR of 1.50 means that the target-language version requires 50\% more tokens than the English source item under the same tokenizer.

\subsection{Effective Context Window}

For a nominal context window $W$, we estimate the language-adjusted effective context window as:

\begin{equation}
\mathrm{ECW}_{\ell,t} = \frac{W}{\overline{\mathrm{TFR}}_{\ell,t}}.
\end{equation}

This metric is not a claim that the model has a different architectural limit for each language. It estimates how much semantically equivalent content can fit before hitting the same token limit.

\subsection{Illustrative API Cost Premium}

For commercial API settings, we estimate the input-token cost premium per 1,000 requests as:

\begin{equation}
\mathrm{Premium}_{\ell} = P \times (\overline{\mathrm{TFR}}_{\ell} - 1) \times \bar{n}_{\mathrm{EN}} \times 1000,
\end{equation}

where $P$ is the input-token price per token and $\bar{n}_{\mathrm{EN}}$ is the mean English token count per item. Dollar values are illustrative and use GPT-4o input pricing of \$2.50 per million tokens as of May 2026~\cite{openaiPricing2026}; these should be updated when provider pricing changes. The primary result is the token ratio.

\section{Results}

\subsection{Tokenization Premiums Across Languages}

Table~\ref{tab:tfr_main} reports mean token fertility ratios across all 120 items. English is the baseline at 1.00 for each tokenizer.

\begin{table*}[t]
\centering
\caption{Token fertility ratio by language and tokenizer. Mean $\pm$ standard deviation is computed across the 120 TEA items. Higher values indicate greater tokenization premium relative to English under the same tokenizer.}
\label{tab:tfr_main}
\begin{tabular}{lccc}
\toprule
\textbf{Language} & \textbf{GPT-4o} & \textbf{Qwen2.5-7B} & \textbf{Mistral-7B} \\
\midrule
English   & 1.00 $\pm$ 0.00 & 1.00 $\pm$ 0.00 & 1.00 $\pm$ 0.00 \\
Arabic    & 1.44 $\pm$ 0.21 & 1.70 $\pm$ 0.27 & 3.86 $\pm$ 0.54 \\
Hindi     & 1.72 $\pm$ 0.31 & 4.86 $\pm$ 0.91 & 5.20 $\pm$ 0.99 \\
Bengali   & 1.56 $\pm$ 0.41 & 4.50 $\pm$ 1.46 & 4.44 $\pm$ 1.45 \\
Tamil     & 2.09 $\pm$ 0.48 & 6.55 $\pm$ 1.80 & 6.57 $\pm$ 1.82 \\
Yoruba    & 2.37 $\pm$ 0.49 & 3.18 $\pm$ 0.63 & 3.33 $\pm$ 0.67 \\
\bottomrule
\end{tabular}
\end{table*}

Three observations stand out. First, every non-English language in the corpus has a tokenization premium under every tokenizer. Second, GPT-4o's tokenizer produces substantially lower premiums than the two open-weight tokenizers for Bengali, Hindi, Tamil, and Arabic, although the penalty remains nontrivial. Third, Yoruba shows the highest GPT-4o premium even though it uses Latin script. This suggests that tokenization inequity is not explained by script family alone. Vocabulary coverage, diacritics, corpus frequency, and tokenizer training composition also matter.
A fourth pattern emerges from the cross-tokenizer comparison: 
the gap between commercial and open-weight models is substantial. 
For Bengali, GPT-4o achieves 1.56$\times$ while Qwen2.5 and 
Mistral reach 4.5$\times$—a nearly threefold improvement.
This suggests that commercial API providers have invested 
substantially more tokenizer capacity in multilingual vocabulary 
coverage than current open-weight models. For learners in 
underserved communities seeking to avoid API costs through 
offline deployment, this creates a trade-off: lower financial 
cost but significantly higher tokenization overhead, which 
manifests as longer sequences, greater memory pressure, and 
reduced effective context.

\begin{figure*}[t]
\centering
\includegraphics[width=0.95\linewidth]{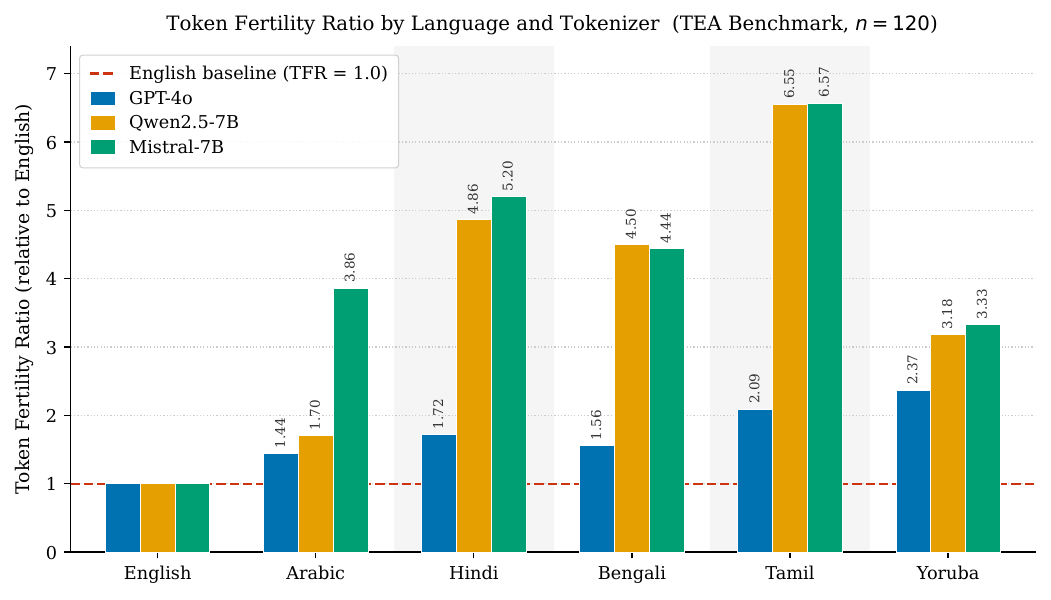}
\caption{Tokenization premiums across the TEA corpus relative to English (baseline = 1.0). Higher values indicate greater token inflation for semantically equivalent content.}
\label{fig:tfr_bar}
\end{figure*}

\subsection{Context Window and Cost Implications}

Table~\ref{tab:cost_context} translates GPT-4o TFR into effective context window estimates for a 128k-token nominal context window and illustrative input-token cost premiums. These estimates use token ratios from the TEA corpus and should be interpreted as operational approximations, not provider-independent constants.

\begin{table}[t]
\centering
\caption{Illustrative GPT-4o context and cost implications. ECW estimates assume a 128k-token nominal context window. Cost estimates use GPT-4o input pricing of \$2.50 per million tokens as of May 2026~\protect\cite{openaiPricing2026}.}

\label{tab:cost_context}
\small
\begin{tabular}{lcc}
\toprule
\textbf{Language} & \textbf{Premium / 1k calls} & \textbf{Effective context} \\
\midrule
English  & \$0.00 & 128,000 (100\%) \\
Arabic   & \$0.06 & 89,148 (70\%) \\
Hindi    & \$0.09 & 74,461 (58\%) \\
Bengali  & \$0.07 & 81,967 (64\%) \\
Tamil    & \$0.14 & 61,121 (48\%) \\
Yoruba   & \$0.18 & 53,951 (42\%) \\
\bottomrule
\end{tabular}
\end{table}

For Bengali, the 1.56$\times$ GPT-4o TFR reduces the amount of semantically equivalent content that fits in a 128k-token context window to roughly 82k English-equivalent tokens. In a tutoring application, this can reduce the amount of prior dialogue, retrieved documentation, or worked examples that can be carried forward. For Yoruba, the effective context estimate falls below half of the nominal window.

\subsection{Bengali Code-Switching Sensitivity}

Table~\ref{tab:sensitivity} reports a Bengali sensitivity analysis for GPT-4o. Removing mixed items does not reduce the aggregate penalty. In Tier 1, the clean-only subset has a higher TFR, indicating that English-retained technical phrases suppress rather than inflate the Bengali tokenization premium.

\begin{table}[t]
\centering
\caption{Bengali TFR sensitivity to technical code-switching under GPT-4o.}
\label{tab:sensitivity}
\small
\begin{tabular}{lccc}
\toprule
\textbf{Subset} & \textbf{T1} & \textbf{T2} & \textbf{T3} \\
\midrule
All items (n=120)  & 1.65 $\pm$ 0.52 & 1.55 $\pm$ 0.36 & 1.50 $\pm$ 0.36 \\
Clean only (n=62)  & 2.11 $\pm$ 0.57 & 1.79 $\pm$ 0.17 & 1.79 $\pm$ 0.10 \\
\bottomrule
\end{tabular}
\end{table}

This result supports treating code-switching as a realistic feature of technical Bengali rather than a data quality failure. It also suggests that tokenization penalties may be understated when English error names and programming identifiers are retained.

\section{Discussion}

\subsection{The Tokenization Premium as Infrastructure Cost}

The TEA results show that tokenization creates a measurable infrastructure cost for underserved language communities. For API-based systems, the cost is visible in token-based billing. For offline or open-weight systems, the cost appears as longer sequences, reduced effective context, and greater memory or latency pressure. These effects are particularly relevant for educational AI because tutoring workflows often require multi-turn context, examples, and explanatory text.

\subsection{Why Script Alone Is Not Enough}

The Yoruba result is important because it prevents a simplistic explanation that only non-Latin scripts are affected. Yoruba's Latin orthography includes tonal and underdot diacritics that may be poorly represented in tokenizer vocabularies. In this corpus, that underrepresentation produces higher GPT-4o fragmentation than Bengali. The broader implication is that tokenizer audits should measure actual language-specific behavior rather than assume fairness from script coverage.

\subsection{Implications for Bengali Educational AI}

For Bengali programming education, tokenization penalties interact with existing resource constraints. Bangla code-generation research has shown the need for dedicated datasets and models \cite{raihan2025tigercoder}. The present audit shows that even when the content exists, the infrastructure used to encode that content can impose additional overhead. This matters for systems targeting low-connectivity learning contexts, where offline deployment may remove API costs but not sequence-length inefficiency.

\subsection{Design Recommendations}

The results suggest four practical recommendations:

\begin{enumerate}
    \item \textbf{Report tokenizer efficiency by language.} Model cards and benchmark papers should include language-level tokenization statistics, especially for underserved languages.
    \item \textbf{Evaluate context-equivalent capacity.} Context window claims should be complemented by language-adjusted estimates when systems are intended for multilingual use.
    \item \textbf{Treat technical domains separately.} Programming education, legal assistance, health guidance, and other terminology-heavy domains should be audited independently because code-switching and domain vocabulary change tokenization behavior.
    \item \textbf{Account for dataset construction inequity.} A 
    100k-token English training corpus becomes 156k tokens 
    in Bengali, 209k in Tamil, and 237k in Yoruba when 
    translated (using GPT-4o TFRs from Table~\ref{tab:tfr_main}). This non-uniform scaling increases storage, 
    transmission bandwidth, and curation costs for underserved 
    languages, creating a structural barrier to equitable 
    benchmark and resource development that operates 
    independently of model quality.
\end{enumerate}

\section{Limitations}

TEA is intentionally small and domain-specific. The 120-item corpus is suitable for a workshop audit but not a comprehensive multilingual tokenizer benchmark. Bengali and Hindi are the fully human-validated target languages in this version; Arabic, Tamil, and Yoruba should be interpreted as exploratory comparisons that motivate broader validation. The code-switching sensitivity analysis is currently restricted to Bengali; extending it to Hindi, Arabic, Tamil, and Yoruba remains a direction for future work, as all target languages retain English programming identifiers and the suppression 
effect observed in Bengali may apply elsewhere.

Dollar-denominated cost estimates are illustrative and should be updated when provider pricing changes. Finally, tokenization efficiency does not by itself measure answer quality. A tokenizer with fewer tokens can still support a weaker model, and a tokenizer with more tokens can still produce useful outputs. TEA measures infrastructure cost, not full pedagogical effectiveness.

\section{Conclusion}

This paper introduced TEA, a reproducible audit for measuring tokenization premiums in underserved-language technical tutoring content. Across a 120-item Python debugging corpus, Bengali requires 1.56$\times$ as many GPT-4o tokens as English and reaches 4.5$\times$ under Qwen2.5 and Mistral tokenizers. Yoruba shows the highest GPT-4o premium despite using Latin script, demonstrating that tokenization inequity is not reducible to script family alone.

These results make a narrow but important point: equitable multilingual AI requires attention to infrastructure layers that operate before model inference. For underserved communities, tokenizer design affects cost, latency, usable context, and local deployment feasibility. Tokenization audits should therefore become part of responsible evaluation for multilingual educational AI systems.

\section*{Data and Code Availability}

The TEA benchmark artifact is publicly available at:

\begin{center}
\url{https://github.com/HeyAvijitRoy/tea-benchmark}
\end{center}

The repository contains the TEA corpus schema, token-counting scripts, aggregate result tables, publication-quality figure-generation code, and reproducibility instructions used in this study. The benchmark includes human-validated Bengali and Hindi translations alongside exploratory Arabic, Tamil, and Yoruba comparison data, consistent with the validation scope described in Section~3.

\section*{Acknowledgments}

This work was supported in part by the Advanced Cyberinfrastructure Coordination Ecosystem: Services \& Support (ACCESS) program through Allocation CIS260616, “Dataset Development and Fine-Tuning of Language Models for Low-Resource Bengali Programming Assistance.” Portions of the dataset processing and benchmark analysis were conducted using ACCESS resources provided through Indiana University Jetstream2. The authors gratefully acknowledge this support.

\bibliographystyle{named}
\bibliography{references}

\end{document}